\documentclass[journal,twoside,web]{ieeecolor}
\usepackage{generic}
\usepackage{amssymb}
\usepackage{hyperref}
\usepackage{amsmath,bm}
\usepackage{algorithm}
\usepackage{algpseudocode}
\usepackage{cite}
\usepackage{subfig}
\usepackage{tabularx}
\usepackage{tabulary}
\usepackage{mathrsfs}
\usepackage{graphicx}
\usepackage{epstopdf}
\usepackage{multirow}
\usepackage{makecell}
\usepackage{booktabs}
\usepackage{url}
\usepackage[dvipsnames]{xcolor}
\hypersetup{hidelinks}
\definecolor{IEEEBlue}{RGB}{0,138,218}
\begin{document}
\bstctlcite{IEEEexample:BSTcontrol}
\title{Correntropy-Based Improper Likelihood Model for Robust Electrophysiological Source Imaging}
\author{Yuanhao~Li,
Badong~Chen,~\IEEEmembership{Senior~Member,~IEEE,}
Zhongxu~Hu,
\\
Keita~Suzuki,
Wenjun~Bai,
Yasuharu~Koike,
and~Okito~Yamashita
\thanks{This work was supported in part by Japan Society for the Promotion of Science (JSPS) KAKENHI under Grants 19H05728, 20H00600, and 23H03433, in part by Innovative Science and Technology Initiative for Security under Grant JPJ004596 ATLA, in part by Moonshot Program 9 under Grant JPMJMS2291, and in part by the National Natural Science Foundation of China under Grants U21A20485 and 62311540022. \emph{(Cor- responding author: Yuanhao Li.)}}
\thanks{Yuanhao Li and Okito Yamashita are with the Center for Advanced Int- elligence Project, RIKEN, Tokyo 103-0027, Japan, and also with the Department of Computational Brain Imaging, ATR Neural Information Analysis Laboratories, Kyoto 619-0237, Japan. (e-mail: yuanhao.li@riken.jp)}
\thanks{Badong Chen is with the Institute of Artificial Intelligence and Roboti- cs, Xi'an Jiaotong University, Xi'an 710049, China.}
\thanks{Zhongxu Hu is with the School of Mechanical Science and Engineering, Huazhong University of Science and Technology, Wuhan 430074, China.}
\thanks{Keita Suzuki and Wenjun Bai are with the Department of Computatio- nal Brain Imaging, ATR Neural Information Analysis Laboratories, Kyoto 619-0237, Japan.}
\thanks{Yasuharu Koike is with the Institute of Innovative Research, Tokyo Ins- titute of Technology, Yokohama 226-8503, Japan.}
\thanks{The code is available at \href{https://sites.google.com/view/liyuanhao/code}{\textcolor{IEEEBlue}{\textit{https://sites.google.com/view/liyuanhao/code}}}.}}

\maketitle

\begin{abstract}
Bayesian learning provides a unified skeleton to solve the electrophysiological source imaging task. From this perspective, existing source imaging algorithms utilize the Gaussian assumption for the observation noise to build the likelihood function for Bayesian inference. However, the electromagnetic measurements of brain activity are usually affected by miscellaneous artifacts, leading to a potentially non-Gaussian distribution for the observation noise. Hence the conventional Gaussian likelihood model is a suboptimal choice for the real-world source imaging task. In this study, we aim to solve this problem by proposing a new likelihood model which is robust with respect to non-Gaussian noises. Motivated by the robust maximum correntropy criterion, we propose a new improper distribution model concerning the noise assumption. This new noise distribution is leveraged to structure a robust likelihood function and integrated with hierarchical prior distributions to estimate source activities by variational inference. In particular, the score matching is adopted to determine the hyperparameters for the improper likelihood model. A comprehensive performance evaluation is performed to compare the proposed noise assumption to the conventional Gaussian model. Simulation results show that, the proposed method can realize more precise source reconstruction by designing known ground-truth. The real-world dataset also demonstrates the superiority of our new method with the visual perception task. This study provides a new backbone for Bayesian source imaging, which would facilitate its application using real-world noisy brain signal.
\end{abstract}


\section{Introduction}
\label{sec:introduction}

\IEEEPARstart{E}{lectrophysiological} source imaging (ESI) aims to reconstruct the intrinsic brain activity by a noninvasive way using magnetoencephalography (MEG) or electroenceph- alography (EEG) measurements. Compared with the functional magnetic resonance imaging (fMRI) modality, ESI can provide the adequate temporal resolution to image the brain dynamics, which has advanced our understanding for cognitive processes and neural bases of mental disorder \cite{baillet2001electromagnetic,he2018electrophysiological}. The key procedure of ESI is to solve the MEG/EEG inverse problem, which refers to estimating the source activity (source space) using the MEG or EEG sensor data (sensor space). To this end, previous works have leveraged two strategies for source activity modeling \cite{he2018electrophysiological}. The first category is the equivalent current dipole (ECD) which models the whole-brain activity by a small number of dipoles. Although ECD will usually lead to an over-determined inverse problem with a unique solution, some drawbacks constrain the applications of ECD, such as the nonlinearity of moving dipole and how to determine the number of dipoles. The second way, which is more popular in recent years, characterizes the source activity as a current density distribution on the cortical surface, using a large number of current dipoles. This is a more realistic assumption, because the MEG/EEG signals primarily originate from the pyramidal cells distributed on the cortical surface \cite{buzsaki2012origin}.

In the distributed source model, MEG/EEG inverse problem is a linear problem, since the locations are fixed for all dipoles. However, it becomes an underdetermined problem that infinite solutions can lead to the same observation, because the number of sources is much larger than that of the MEG/EEG channels. In the past three decades, researchers have investigated various strategies to regularize the solution, such as the minimum norm estimation (MNE) \cite{hamalainen1994interpreting}, weighted MNE (wMNE) \cite{wang1992magnetic}, LORETA \cite{pascual1994low}, standardized LORETA \cite{pascual2002standardized}, beamforming \cite{van1997localization}, scanning \cite{mosher1992multiple}, Bayesian algorithms \cite{sato2004hierarchical,wipf2009unified,wipf2010robust,mohseni2014non,costa2017bayesian,pirondini2017computationally,cai2018hierarchical,liu2019bayesian,cai2019robust,suzuki2021meg,cai2021robust,cai2022bayesian,hashemi2022joint,cai2023bayesian}, etc. Remarkably, almost every source imaging algorithm for the distributed source model can be unified into the Bayesian learning framework with different regularization terms and optimization approaches \cite{wipf2009unified}. Recent advances for Bayesian source imaging have considered source correlation \cite{wipf2010robust,liu2019bayesian,cai2023bayesian}, source in non-Gaussian distribution \cite{mohseni2014non,costa2017bayesian}, simultaneous estimation for both source covariance and noise covariance \cite{pirondini2017computationally,cai2021robust,cai2022bayesian}, spatial extents for source distribution \cite{cai2018hierarchical,cai2019robust}, fMRI prior information of meta analysis \cite{suzuki2021meg}, full structure for noise covariance matrix \cite{hashemi2022joint}, and so on.

Despite these encouraging improvements achieved by recent advances, one significant question has not been fully addressed for every Bayesian source estimation algorithm: How to model the sensor noises? All the Bayesian source imaging algorithms, to the best of our knowledge, assume that the observation noise follows a multivariate Gaussian distribution. Nevertheless, this overly idealistic Gaussian assumption may not characterize the real-world noises adequately. Previous studies pointed out that, the recording noise in MEG/EEG signals is a complex mixture, comprising system-related measurement noises, environmental noises, and physiological artifacts \cite{gross2013good,ball2009signal}. In particular, the sporadic physiological artifacts, including head movement and eye blink, usually result in an larger amplitude than the normal recordings. This suggests that the true noise distribution should be a heavy-tailed non-Gaussian distribution. Though one could denoise the MEG/EEG signal to some extent by preprocessing \cite{mutanen2018automatic}, it is difficult to guarantee that all the non-Gaussian noises could be totally removed. As a result, existing Bayesian source imaging algorithms may be limited by the unrealistic Gaussian noise assumption.

In this study, we aim to address the potentially non-Gaussian observation noises for the Bayesian MEG/EEG source imaging task by proposing a robust noise modeling method. To this end, we introduce the Maximum Correntropy Criterion (MCC) into the Bayesian learning framework, which is utilized to structure the noise distribution and the likelihood function, replacing the conventional Gaussian model for the sensor noise. MCC is one important approach in the information theoretic learning (ITL) \cite{principe2010information} and has competently improved the robustness with respect to non-Gaussian noise in many signal processing and machine learning tasks \cite{liu2007correntropy,he2010maximum,chen2017maximum,ma2018bias,li2023partial,li2023correntropy,li2023adaptive}. To characterize the sensor noise with the robust MCC, the crucial step is to derive the corresponding noise assumption in an explicit form, which will constitute the basis for the likelihood function and the processes for Bayesian inference. Although our previous studies have investigated this derivation \cite{li2023correntropy,li2023adaptive}, they have yet to provide an adequate form for the noise distribution. Moreover, for the previous form, the hyperparameter has to be selected by cross-validation, whereas one has no access to a validation set in the unsupervised source imaging task. Hence, it would be difficult to choose the proper hyperparameter. We aim to solve these obstacles in the present study to realize a better integration of MCC into the Bayesian learning regime. The main contributions of this paper are listed as follows:

\begin{itemize}
	\item[1.]
	A novel distribution is derived which is the inherent noise assumption for MCC. Compared to the preliminary forms \cite{li2023correntropy,li2023adaptive}, this new distribution can better characterize the noise properties.
	\item[2.]
	This novel noise model is an improper distribution, where the integral of the probability density function (PDF) over the entire support is infinite. Hence, the novel distribution cannot be normalized, and traditional ways for estimating the distribution parameters will fail, such as the maximum likelihood estimation. To address this, we utilize the score matching technique \cite{hyvarinen2005estimation}, which provides an effective tool for non-normalized models. Through this method, a novel approach for selecting the proper hyperparameter of MCC is proposed without the need for a validation set.
	\item[3.]
	The novel noise assumption is leveraged to derive a robust likelihood function, which is integrated into the Bayesian source imaging framework. Variational inference method is employed to calculate the maximum a posteriori (MAP) estimation for the source distribution.
	\item[4.]
	Concerning the performance evaluation, this work utilizes a simulation study and a real-world dataset to fully assess the efficacy for the proposed source estimation algorithm.
\end{itemize}

The code for the simulation study is available at \href{https://sites.google.com/view/liyuanhao/code}{\textcolor{IEEEBlue}{\textit{https://sites. google.com/view/liyuanhao/code}}}.

\section{Method}
\label{sec:method}

This section first presents a brief review for Bayesian source imaging and MCC. Then, we derive a novel MCC-based noise assumption, and discuss its crucial properties. This assumption is utilized to structure a robust likelihood function. Finally, we propose a robust Bayesian source imaging algorithm to address the non-Gaussian observation noise.

\subsection{Bayesian Electrophysiological Source Imaging}
\label{sec:esi}

The distributed source model for the ESI task supposes that MEG/EEG signal is generated by the following forward model
\begin{equation}
	\label{equ:esi1}
	\boldsymbol{b}=\boldsymbol{Gj}+\boldsymbol{\varepsilon}
\end{equation}
where $\boldsymbol{j}=(j_1,...,j_N)^\intercal$ denotes neural activities of $N$ vertices on the cortical surface, $\boldsymbol{b}=(b_1,...,b_M)^\intercal$ represents the sensor data with $M$ MEG/EEG channels. $\boldsymbol{G}\in\mathbb{R}^{M\times N}$ is the leadfield matrix, expressing the conduction from the source space to the sensor space. $\boldsymbol{\varepsilon}=(\varepsilon_1,...,\varepsilon_M)^\intercal$ denotes the observation noise for each MEG/EEG channel. If one considers time series data, the forward model (\ref{equ:esi1}) can be written using the following form:
\begin{equation}
	\label{equ:esi2}
	\boldsymbol{B}=\boldsymbol{GJ}+\boldsymbol{E}
\end{equation}
where $\boldsymbol{J}=(\boldsymbol{j}_1,...,\boldsymbol{j}_T)\in\mathbb{R}^{N\times T}$, $\boldsymbol{B}=(\boldsymbol{b}_1,...,\boldsymbol{b}_T)\in\mathbb{R}^{M\times T}$, and $\boldsymbol{E}=(\boldsymbol{\varepsilon}_1,...,\boldsymbol{\varepsilon}_T)\in\mathbb{R}^{M\times T}$ consist of $T$ samples. Because $N\gg M$ for the distributed source model, solving (\ref{equ:esi2}) becomes an ill-posed problem, which requires a regularization on source estimation.

The Bayesian learning framework provides an effective tool for this task, which will facilitate the incorporation of different constraints on the source data. In this paper we utilize a similar formulation as our previous work \cite{suzuki2021meg}. To begin with, the noise at $t$-th time point, denoted by $\boldsymbol{\varepsilon}_t$, is assumed to follow a multi- variate Gaussian distribution $\mathcal{N}(\boldsymbol{0},(\beta\boldsymbol{\varPhi})^{-1})$, in which $\boldsymbol{\varPhi}$ is the normalized noise precision matrix satisfying $Tr(\boldsymbol{\varPhi})=M$, and is typically measured by the pre-stimulus period. $\beta$ is a scaling parameter. Based on the forward model (\ref{equ:esi1}), the PDF of sensor data at $t$-th time point is
\begin{equation}
	\label{equ:esi3}
	P(\boldsymbol{b}_t\mid\boldsymbol{j}_t,\beta)=\mathcal{N}(\boldsymbol{b}_t\mid\boldsymbol{Gj}_t,(\beta\boldsymbol{\varPhi})^{-1})
\end{equation}
By supposing the independence between every time point, one could obtain the likelihood function for the time series matrix:
\begin{equation}
	\label{equ:esi4}
	P(\boldsymbol{B}\mid\boldsymbol{J},\beta)=\prod_{t=1}^{T}\mathcal{N}(\boldsymbol{b}_t\mid\boldsymbol{Gj}_t,(\beta\boldsymbol{\varPhi})^{-1})
\end{equation}
From a Bayesian viewpoint, the constraint regarding the source activity can be expressed by a specific prior distribution on $\boldsymbol{J}$:
\begin{equation}
	\label{equ:esi5}
	P_0(\boldsymbol{J}\mid\boldsymbol{\varLambda})=\prod_{t=1}^{T}P_0(\boldsymbol{j}_t\mid \boldsymbol{\varLambda})
\end{equation}
with a hyperparameter $\boldsymbol{\varLambda}$. Previous works have mainly focused on designing better $P_0(\boldsymbol{J}\mid\boldsymbol{\varLambda})$ \cite{wipf2009unified,wipf2010robust,mohseni2014non,costa2017bayesian,cai2018hierarchical,cai2019robust,suzuki2021meg,cai2023bayesian}, or proposing a more efficient Bayesian inference method \cite{pirondini2017computationally,cai2021robust,cai2022bayesian}. By formulating the likelihood function and the prior distribution, the source activity $\boldsymbol{J}$ is estimated by the posterior distribution using Bayes' theorem:
\begin{equation}
	\label{equ:esi6}
	P(\boldsymbol{J}\mid\boldsymbol{B})=\frac{P(\boldsymbol{B}\mid\boldsymbol{J},\beta)P_0(\boldsymbol{J}\mid\boldsymbol{\varLambda})}{P(\boldsymbol{B})}
\end{equation}

In this paper, we utilize the hierarchical variational Bayesian (hVB) model in \cite{sato2004hierarchical} as the baseline, which could appropriately incorporate the fMRI prior information. hVB assumes the prior distribution for $\boldsymbol{J}$ by the following hierarchical manner:
\begin{equation}
	\label{equ:hvb1}
	P_0(\boldsymbol{J}\mid\boldsymbol{A})=\prod_{t=1}^{T}\mathcal{N}(\boldsymbol{j}_t\mid\boldsymbol{0},\boldsymbol{A}^{-1})
\end{equation}
\begin{equation}
	\label{equ:hvb2}
	P_0(\boldsymbol{A})=\prod_{n=1}^{N}P_0(a_n)=\prod_{n=1}^{N}\varGamma(a_n\mid a_{n0},\gamma_0)
\end{equation}
\begin{equation}
	\label{equ:hvb3}
	P_0(\beta)=1/\beta
\end{equation}
in which $a_n$ denotes the inverse variance for the source activity on the $n$-th vertex and $\boldsymbol{A}=diag(a_1,...,a_N)\in\mathbb{R}^{N\times N}$ denotes the diagonal precision matrix for the source signal. The inverse variance $a_n$ follows the Gamma distribution $\varGamma(a_n\mid a_{n0},\gamma_0)=\frac{(\gamma_0/a_{n0})^{\gamma_0}}{\varGamma(\gamma_0)}a_n^{\gamma_0-1}\exp(-a_n\gamma_0/a_{n0})$, with mean $a_{n0}$ and degree of freedom $\gamma_0$, and $\varGamma(\gamma_0)$ is the Gamma function. hVB utilizes fMRI statistics to determine $a_{n0}$. The noise scaling parameter $\beta$ is presumed with the non-informative prior distribution (\ref{equ:hvb3}). The joint posterior distribution is calculated by
\begin{equation}
	\label{equ:hvb4}
	P(\boldsymbol{J},\boldsymbol{A},\beta\mid\boldsymbol{B})=\frac{P(\boldsymbol{B}\mid\boldsymbol{J},\beta)P_0(\boldsymbol{J}\mid\boldsymbol{A})P_0(\boldsymbol{A})P_0(\beta)}{P(\boldsymbol{B})}
\end{equation}
Since it would be difficult to analytically compute the marginal likelihood $P(\boldsymbol{B})$, the variational Bayesian method is leveraged to calculate the MAP estimations for all the random variables.

\subsection{MCC-Based Noise Distribution Model}
\label{sec:model}

\subsubsection{Maximum Correntropy Criterion}~
\label{sec:mcc}

Correntropy was first developed as a generalized correlation function for nonlinear space, which has been further employed as a measure of similarity between two random variables \cite{liu2007correntropy}. Given two random variables $x$ and $y$, the correntropy is defined by
\begin{equation}
	\label{equ:corr1}
	V(x,y)\triangleq\mathbb{E}_{P(x,y)}[k(x,y)]
\end{equation}
which represents the expectation of the kernel function $k(x,y)$ with respect to the joint distribution $P(x,y)$. The most popular kernel function for correntropy is the Gaussian kernel function:
\begin{equation}
	\label{equ:corr2}
	k_{h_c}(x,y)\triangleq\exp(-\frac{(x-y)^2}{2h_c})
\end{equation}
in which $h_c>0$ denotes the kernel bandwidth for correntropy. Compared to the long-established second-order statistics, using correntropy as the similarity measure provides several benefits: 1) correntropy is a local similarity measure which is insensitive to the outliers; 2) it contains all the even-order moments which are effective for the non-Gaussian data analysis; 3) the weights for different values are determined by the kernel bandwidth $h_c$.

In supervised learning, one could maximize the correntropy between the prediction and the desired target, called Maximum Correntropy Criterion (MCC). Let $x$ and $y$ represent the model prediction and the target, respectively. MCC could be achieved by maximizing the empirical estimation of correntropy $V(x,y)$ using $L$ samples:
\begin{equation}
	\label{equ:corr3}
	\max \frac{1}{L}\sum_{i=1}^{L}\exp(-\frac{(x_i-y_i)^2}{2h_c})
\end{equation}
By defining the residual term $e_i=x_i-y_i$, MCC is also written by $\max\frac{1}{L}\sum_{i=1}^{L}\exp(-\frac{1}{2}e_i^2/h_c)$. Hence, the loss function with respect to the error for MCC is $\mathcal{L}_{MCC}(e)=-\exp(-\frac{1}{2}e^2/h_c)$.

\subsubsection{Inherent Noise Assumption in MCC}~
\label{sec:mcc_noise}

Given an arbitrary loss function $\mathcal{L}(e)$, the probability model of $P(e)=\exp(-\mathcal{L}(e))$ would provide the likelihood function that leads to the same solution of minimizing the loss function $\mathcal{L}(e)$ \cite{jewson2022general}, e.g., Gaussian likelihood and the mean square error (MSE) loss. Although this may lead to an improper distribution model, where the integral for $P(e)$ is infinite, and $P(e)$ cannot be normalized, this transform provides a method to explore the inherent noise assumption within a loss function. Our previous study \cite{li2023adaptive} utilized this method to expose the noise assumption for MCC, yielding the following distribution:
\begin{equation}
	\label{equ:corrmodel1}
	\mathcal{C}_{old}(e\mid0,h_c)=\exp(\exp(-\frac{e^2}{2h_c}))
\end{equation}
which is a zero-mean improper distribution model. The kernel bandwidth $h_c$ acts as the scaling parameter for this distribution. However, this old formulation has several problems. First, one could know from \cite{liu2007correntropy} that, an infinite kernel bandwidth $h_c\rightarrow+\infty$ will cause MCC to degenerate into the MSE loss function, whereas $\mathcal{C}_{old}$ will not degenerate into the Gaussian distribution, which appears to be a conflict. Since $\mathcal{C}_{old}$ cannot approximate the Gaussian distribution for any $h_c$, it would result in a biased solution in the presence of Gaussian noises, implying a limited range of applicability. Moreover, the hyperparameter $h_c$ is used to reflect the characteristics of the out-of-distribution noise (i.e. outlier). Thus, $\mathcal{C}_{old}$ cannot define the property of in-distribution noise, which is inadequate for Bayesian source estimation task.

\begin{figure}[t!]
	\centering
	\includegraphics[width=1\columnwidth]{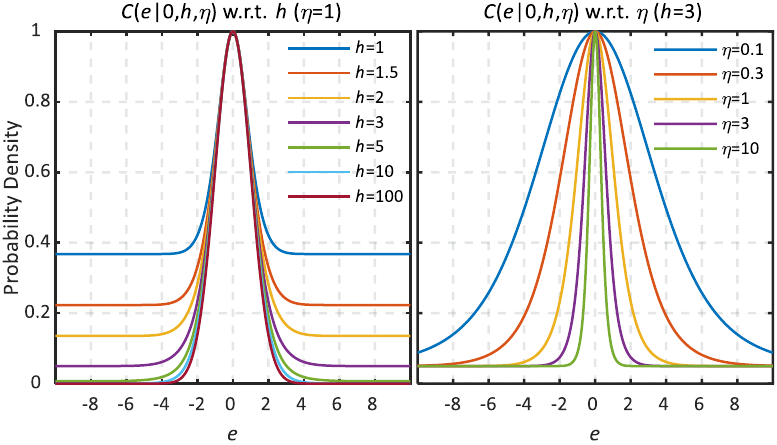}
	\caption{MCC-based noise assumption $\mathcal{C}(e|0,h,\eta)$ with respect to different $h$ values (left panel, $\eta=1$), and different $\eta$ values (right panel, $h=3$).}
	\label{FigMccCurve}
\end{figure}

In this work, we desire to propose a better distribution model motivated by MCC to solve the above-mentioned issues. In this regard, the following new noise distribution model is proposed:
\begin{equation}
	\label{equ:corrmodel2}
	\mathcal{C}(e\mid0,h,\eta)=\exp\left(h\cdot\exp(-\frac{\eta e^2}{2h})-h\right)
\end{equation}
which is a zero-centered distribution with two hyperparameters $h,\eta>0$. $h$ is named robustness parameter, which could reflect the level of outliers. $\eta$ is named dispersion parameter, defining the distribution for regular noises. Fig. \ref{FigMccCurve} shows some examples of $\mathcal{C}(e\mid0,h,\eta)$ with different $h$ and $\eta$. One could observe that, this novel form is also an improper distribution model, because $\lim_{e\rightarrow\infty}\mathcal{C}(e\mid0,h,\eta)>0$ and therefore its integral is infinity. One can also observe the roles of $h$ and $\eta$ for this distribution. Specifically, $h$ controls the out-of-distribution tail, which plays a vital role in handling outliers \cite{li2023adaptive}. A smaller $h$ value results in a heavier tail that is suitable for a larger fraction of outliers. On the other hand, $\eta$ describes the distribution of regular noise, where a smaller $\eta$ value implies a greater dispersion. Thus, the in-distribution noise is well described by the novel model (\ref{equ:corrmodel2}). Furthermore, we would like to discuss two essential properties for this novel distribution as follows:

\textit{Property 1:} The maximum likelihood estimation that utilizes the noise assumption $\mathcal{C}(e\mid0,h,\eta)$ is equivalent to the original MCC.

\textit{Proof:} By assuming the independence between each sample, the likelihood function with $L$ error samples $\{e_i\}_{i=1}^L$ is:
\begin{equation}
	\label{equ:corrmodel3}
	\prod_{i=1}^{L}\mathcal{C}(e_i\mid0,h,\eta)=\prod_{i=1}^{L}\exp\left(h\cdot\exp(-\frac{\eta e_i^2}{2h})-h\right)
\end{equation}

If one applies the logarithm function to the outer exponential function, the maximum likelihood estimation is:
\begin{equation}
	\begin{split}
		\label{equ:corrmodel4}
		\max&\prod_{i=1}^{L}\exp\left(h\cdot\exp(-\frac{\eta e_i^2}{2h})-h\right) \; \\
		\Longleftrightarrow\max&\sum_{i=1}^{L}\left(h\cdot\exp(-\frac{\eta e_i^2}{2h})-h\right) \; \\
		\Longleftrightarrow\max&\frac{1}{L}\sum_{i=1}^{L}\left(\exp(-\frac{\eta e_i^2}{2h})\right) \; \\
	\end{split}
\end{equation}
where the first equation is achieved by the logarithm function, and the second equation holds because $h$ and $L$ are both fixed constants. One observes that, (\ref{equ:corrmodel4}) will be equal to the original MCC (\ref{equ:corr3}) if $h_c=h/\eta$. As a result, all the advantages of MCC could be well inherited by this novel noise distribution model.

\textit{Property 2:} The proposed $\mathcal{C}(e\mid0,h,\eta)$ will degenerate into a Gaussian distribution when $h\rightarrow+\infty$.

\textit{Proof:} By doing a Taylor expansion on the inner exponential function, one has:
\begin{equation}
	\begin{split}
		\label{equ:corrmodel5}
		&\lim_{h\rightarrow+\infty}\exp\left(h\cdot\exp(-\frac{\eta e^2}{2h})-h\right) \; \\
		=&\lim_{h\rightarrow+\infty}\exp\left(\sum_{n=0}^{\infty}(\frac{(-1)^n}{2^nn!}\cdot\frac{\eta^ne^{2n}}{h^{n-1}})-h\right) \; \\
		=&\lim_{h\rightarrow+\infty}\exp\left(h-\frac{1}{2}\eta e^2+\frac{1}{8}\eta^2\frac{e^4}{h}-\cdots-h\right) \; \\
		=&\exp(-\frac{1}{2}\eta e^2) \propto\mathcal{N}(e\mid0,\eta^{-1})\; \\
	\end{split}
\end{equation}
which is a zero-mean Gaussian distribution, and the dispersion parameter $\eta$ becomes the inverse variance. Thus, $\mathcal{C}(e\mid0,h,\eta)$ can be viewed as a generalization for the Gaussian distribution.

Through the above discussion, one can observe that, the new $\mathcal{C}(e\mid0,h,\eta)$ addresses all the problems in $\mathcal{C}_{old}$. First, the new form resolves the conflict of degeneration from $\mathcal{C}$ to Gaussian distribution, as compared to that from MCC to MSE. This also implies that, $\mathcal{C}(e\mid0,h,\eta)$ is suitable to handle Gaussian noise. In addition, $\mathcal{C}(e\mid0,h,\eta)$ has an extra parameter $\eta$ to describe the in-distribution noise.

\subsubsection{Determination of Hyperparameter}~
\label{sec:mcc_hyper}

Two hyperparameters need to be determined in the proposed $\mathcal{C}(e\mid0,h,\eta)$, characterizing the level of outliers and dispersion of regular noises, respectively. However, because $\mathcal{C}(e\mid0,h,\eta)$ is an improper model, traditional methods for determining the distribution parameters will be inapplicable, such as maximum likelihood function with respect to the distribution parameters using a given dataset, or optimizing the distribution parameters with variational inference. Not to mention, in the unsupervised source imaging scenario, one cannot obtain a validation dataset with desired target, and therefore cross-validation is inadequate for selecting the hyperparameters.

To address this obstacle, we employ the score matching \cite{hyvarinen2005estimation} method to choose the hyperparameters $h$ and $\eta$. To be specific, the $\mathcal{H}$-score for the proposed $\mathcal{C}(e\mid0,h,\eta)$ is define as follows:
\begin{equation}
	\label{equ:corrmodel6}
	\mathcal{H}_{\mathcal{C}}(e)\triangleq2\frac{\partial^2}{\partial e^2}\log\mathcal{C}(e\mid0,h,\eta)+\left(\frac{\partial}{\partial e}\log\mathcal{C}(e\mid0,h,\eta)\right)^2
\end{equation}
One could observe that, $\mathcal{H}$-score is only related to the first and second derivatives of $\log\mathcal{C}(e\mid0,h,\eta)$. As a result, the infinite normalizing constant could be circumvented. With $L$ samples $\{e_i\}_{i=1}^L$, the optimal hyperparameters $h^*$ and $\eta^*$ are given by:
\begin{equation}
	\label{equ:corrmodel7}
	(h^*,\eta^*)=arg\min_{h,\eta}\frac{1}{L}\sum_{i=1}^{L}\mathcal{H}_{\mathcal{C}}(e_i)
\end{equation}
which could be solved by alternately optimizing $h$ and $\eta$ using gradient method. The score matching in essence minimizes the Fisher’s divergence between $\mathcal{C}(e\mid0,h,\eta)$ and the ground-truth data-generating distribution \cite{hyvarinen2005estimation}.

Despite the efficiency of score matching for non-normalized models, another critical problem is that, one could never obtain the ground-truth noise set. In this regard, we leverage a similar empirical method as in our previous study \cite{li2021restricted}, which employs the baseline algorithm to obtain a rough estimate for the noise. Specifically, we first use the conventional Gaussian distribution as the noise assumption to optimize the model parameter. Then we compute the residuals. If the ground-truth noise distribution is Gaussian, the Gaussian likelihood will provide the unbiased solution, and the residuals will exhibit a Gaussian distribution. Thus, if we estimate $h$ and $\eta$ using the residuals, $\mathcal{C}(e\mid0,h,\eta)$ will degenerate into a Gaussian distribution. Conversely, if the noise follows a heavy-tailed distribution, the residuals obtained by Gaussian likelihood will reveal a non-Gaussian distribution. Thus, using the residuals of the Gaussian likelihood model for score matching can provide an approximate estimation for the ground-truth noise distribution.

\subsection{Robust Source Imaging with MCC}
\label{sec:mcc_bayes}

This subsection aims to derive the source imaging algorithm using the novel noise assumption $\mathcal{C}(e\mid0,h,\eta)$ and hierarchical prior distributions defined by the hVB algorithm. First, one can assume that the hyperparameters $h$ and $\eta$ have been determined for each sensor channel individually, yielding $(h_1,...,h_M)$ and $(\eta_1,...,\eta_M)$. Hence, the noise for $m$-th channel $\varepsilon_m$ follows the distribution $\mathcal{C}(\varepsilon_m\mid0,h_m,\eta_m)$, and the observation $b_m$ follows $\mathcal{C}(b_m\mid(\boldsymbol{Gj})_m,h_m,\eta_m)$, in which $(\boldsymbol{Gj})_m$ expresses the $m$-th element of $\boldsymbol{Gj}$. By assuming the independence between each sensor channel and each time sampling, the likelihood function is written by:
\begin{equation}
	\begin{split}
		\label{equ:chvb1}
		&P(\boldsymbol{B}\mid\boldsymbol{J})=\prod_{t=1}^{T}\prod_{m=1}^{M}\mathcal{C}(\boldsymbol{B}_{m,t}\mid(\boldsymbol{GJ})_{m,t},h_m,\eta_m)\; \\
		&=\prod_{t=1}^{T}\prod_{m=1}^{M}\exp\left(h_m\cdot\exp(-\frac{\eta_m \boldsymbol{E}_{m,t}^2}{2h_m})-h_m\right)\; \\
	\end{split}
\end{equation}
in which the subscript $(m,t)$ indicates the index of the element in each matrix, and $\boldsymbol{E}=\boldsymbol{B}-\boldsymbol{GJ}$ denotes the reconstruction error. However, one could observe that, this likelihood function  is non-conjugate with the prior distribution defined in the hVB algorithm. Therefore, to obtain the MAP estimation, we utilize the variational inference \cite{gelman1995bayesian} which minimizes the free energy:
\begin{equation}
	\label{equ:chvb2}
	\min-\mathbb{E}_{Q_{\boldsymbol{J}}(\boldsymbol{J})Q_{\boldsymbol{A}}(\boldsymbol{A})}\left[\log\frac{P(\boldsymbol{B},\boldsymbol{J},\boldsymbol{A})}{Q_{\boldsymbol{J}}(\boldsymbol{J})Q_{\boldsymbol{A}}(\boldsymbol{A})}\right]
\end{equation}
in which $Q_{\boldsymbol{J}}(\boldsymbol{J})$ and $Q_{\boldsymbol{A}}(\boldsymbol{A})$ represent the proxy distributions of $\boldsymbol{J}$ and $\boldsymbol{A}$, respectively, to estimate the posterior distribution. The free energy minimization could be achieved by alternately updating $Q_{\boldsymbol{J}}(\boldsymbol{J})$ and $Q_{\boldsymbol{A}}(\boldsymbol{A})$:
\begin{equation}
	\begin{split}
		\label{equ:chvb3}
		&\log Q_{\boldsymbol{J}}(\boldsymbol{J})=\mathbb{E}_{Q_{\boldsymbol{A}}(\boldsymbol{A})}\left[\log P(\boldsymbol{B},\boldsymbol{J},\boldsymbol{A})\right]+const.\; \\
		&\log Q_{\boldsymbol{A}}(\boldsymbol{A})=\mathbb{E}_{Q_{\boldsymbol{J}}(\boldsymbol{J})}\left[\log P(\boldsymbol{B},\boldsymbol{J},\boldsymbol{A})\right]+const.\; \\
	\end{split}
\end{equation}
Substituting $P(\boldsymbol{B},\boldsymbol{J},\boldsymbol{A})=P(\boldsymbol{B}\mid\boldsymbol{J})P_0(\boldsymbol{J}\mid\boldsymbol{A})P_0(\boldsymbol{A})$ gives:
\begin{equation}
	\begin{split}
		\label{equ:chvb4}
		\log Q_{\boldsymbol{J}}(\boldsymbol{J})=&\sum_{t=1}^{T}\sum_{m=1}^{M}h_m\cdot\exp(-\frac{\eta_m \boldsymbol{E}_{m,t}^2}{2h_m})\; \\
		&-\frac{1}{2}\sum_{t=1}^{T}\boldsymbol{j}_t^\intercal\mathbb{E}_{Q_{\boldsymbol{A}}(\boldsymbol{A})}\left[\boldsymbol{A}\right]\boldsymbol{j}_t+const.\; \\
	\end{split}
\end{equation}
\begin{equation}
	\begin{split}
		\label{equ:chvb5}
		\log Q_{\boldsymbol{A}}(\boldsymbol{A})=&\sum_{n=1}^{N}\left((\gamma_0+\frac{T}{2}-1)\log a_n-\frac{\gamma_0}{a_{n0}}a_n\right)\; \\
		&-\frac{1}{2}\sum_{n=1}^{N}\sum_{t=1}^{T}\mathbb{E}_{Q_{\boldsymbol{J}}(\boldsymbol{J})}\left[\boldsymbol{J}_{n,t}^2\right]a_n+const.\; \\
	\end{split}
\end{equation}

Regarding the expectation of squared source $\mathbb{E}_{Q_{\boldsymbol{J}}(\boldsymbol{J})}\left[\boldsymbol{J}_{n,t}^2\right]$, we use Laplacian approximation for $\log Q_{\boldsymbol{J}}(\boldsymbol{J})$, so that $Q_{\boldsymbol{J}}(\boldsymbol{J})$ will follow a Gaussian distribution. Here, we treat $\log Q_{\boldsymbol{J}}(\boldsymbol{J})$ for each time separately using $\log Q_{\boldsymbol{J}}(\boldsymbol{J})=\sum_{t=1}^{T}\log Q_{\boldsymbol{J}}(\boldsymbol{j}_t)$:
\begin{equation}
	\label{equ:chvb6}
	\log Q_{\boldsymbol{J}}(\boldsymbol{j}_t)=\sum_{m=1}^{M}h_m\exp(-\frac{\eta_m \boldsymbol{E}_{m,t}^2}{2h_m})-\frac{1}{2}\boldsymbol{j}_t^\intercal\mathbb{E}_{Q_{\boldsymbol{A}}(\boldsymbol{A})}\left[\boldsymbol{A}\right]\boldsymbol{j}_t
\end{equation}
Each $\log Q_{\boldsymbol{J}}(\boldsymbol{j}_t)$ is then applied with Laplacian approximation
\begin{equation}
	\label{equ:chvb7}
	\log Q_{\boldsymbol{J}}(\boldsymbol{j}_t)\approx \log Q_{\boldsymbol{J}}(\boldsymbol{j}_t^*)-\frac{1}{2}(\boldsymbol{j}_t-\boldsymbol{j}_t^*)^\intercal\mathbf{H}(\boldsymbol{j}_t^*)(\boldsymbol{j}_t-\boldsymbol{j}_t^*)
\end{equation}
in which $\boldsymbol{j}_t^*$ is the maximum point for $\log Q_{\boldsymbol{J}}(\boldsymbol{j}_t)$, and $\mathbf{H}(\boldsymbol{j}_t^*)$ denotes the negative Hessian matrix of $\log Q_{\boldsymbol{J}}(\boldsymbol{j}_t)$ at $\boldsymbol{j}_t^*$. Thus, $Q_{\boldsymbol{J}}(\boldsymbol{j}_t)$ is approximated by the following Gaussian distribution
\begin{equation}
	\label{equ:chvb8}
	Q_{\boldsymbol{J}}(\boldsymbol{j}_t)\approx \mathcal{N}(\boldsymbol{j}_t\mid\boldsymbol{j}_t^*,\mathbf{H}(\boldsymbol{j}_t^*)^{-1})
\end{equation}
As a result, the expectation of squared source is computed by:
\begin{equation}
	\label{equ:chvb9}
	\mathbb{E}_{Q_{\boldsymbol{J}}(\boldsymbol{j}_t)}\left[\boldsymbol{J}_{n,t}^2\right]=(\boldsymbol{j}_t^*)_n^2+\left[\mathbf{H}(\boldsymbol{j}_t^*)^{-1}\right]_{n,n}
\end{equation}
where $(\boldsymbol{j}_t^*)_n$ is the $n$-th entry of $\boldsymbol{j}_t^*$, and $\left[\mathbf{H}(\boldsymbol{j}_t^*)^{-1}\right]_{n,n}$ denotes the $n$-th diagonal component of $\mathbf{H}(\boldsymbol{j}_t^*)^{-1}$. To calculate $\boldsymbol{j}_t^*$, we could set the gradient of $\log Q_{\boldsymbol{J}}(\boldsymbol{j}_t)$ with respect to $\boldsymbol{j}_t$ to zero, which would lead to the following fixed-point update rule \cite{chen2015convergence}:
\begin{equation}
	\label{equ:chvb10}
	\boldsymbol{j}_t=(\boldsymbol{G}^\intercal \boldsymbol{\varPsi}\boldsymbol{G}+\mathbb{E}_{Q_{\boldsymbol{A}}(\boldsymbol{A})}\left[\boldsymbol{A}\right])^{-1}\boldsymbol{G}^\intercal\boldsymbol{\varPsi}\boldsymbol{b}_t
\end{equation}
where $\boldsymbol{\varPsi}\in\mathbb{R}^{M\times M}$ is a diagonal matrix, and the $m$-th element is $\boldsymbol{\varPsi}_{m,m}=\eta_m\cdot\exp(-\eta_m \boldsymbol{E}_{m,t}^2/2h_m)$.

After obtaining $\boldsymbol{j}_t^*$ for each $\log Q_{\boldsymbol{J}}(\boldsymbol{j}_t)$ and calculating the expectation of the squared source $\mathbb{E}_{Q_{\boldsymbol{J}}(\boldsymbol{J})}\left[\boldsymbol{J}_{n,t}^2\right]$ utilizing (\ref{equ:chvb9}), one observes that, $Q_{\boldsymbol{A}}(\boldsymbol{A})$ could be formed into the following Gamma distribution through (\ref{equ:chvb5}):
\begin{equation}
	\label{equ:chvb11}
	Q_{\boldsymbol{A}}(\boldsymbol{A})=\prod_{n=1}^{N}Q_{\boldsymbol{A}}(a_n)=\prod_{n=1}^{N}\varGamma(a_n\mid \bar{a}_n,\bar{\gamma})
\end{equation}
where $\bar{\gamma}=\gamma_0+T/2$, and the expectation $\bar{a}_n$ is optimized by:
\begin{equation}
	\label{equ:chvb12}
	\bar{a}_n=\frac{\gamma_0+T/2}{\gamma_0/a_{n0}+\frac{1}{2}\sum_{t=1}^{T}\mathbb{E}_{Q_{\boldsymbol{J}}(\boldsymbol{J})}\left[\boldsymbol{J}_{n,t}^2\right]}
\end{equation}
With defining the prior weight $w\triangleq\gamma_0/(\gamma_0+T/2)$, the update rule for the expectation $\bar{a}_n$ could be also represented by \cite{suzuki2021meg}:
\begin{equation}
	\label{equ:chvb13}
	\bar{a}_n^{-1}=w\cdot a_{n0}^{-1}+(1-w)\cdot\frac{1}{T}\sum_{t=1}^{T}\mathbb{E}_{Q_{\boldsymbol{J}}(\boldsymbol{J})}\left[\boldsymbol{J}_{n,t}^2\right]
\end{equation}
for which the prior weight $w$ controls the degree of confidence for the fMRI prior information $a_{n0}$. Thus, the updated $\bar{a}_n$ can be substituted back into $\mathbb{E}_{Q_{\boldsymbol{A}}(\boldsymbol{A})}\left[\boldsymbol{A}\right]$ in (\ref{equ:chvb4}), which completes the variational inference processes by alternating the iterations.

The reformulated hVB source imaging algorithm leveraging the correntropy-based noise assumption $\mathcal{C}(e\mid0,h,\eta)$ is named as ChVB algorithm, as summarized in Algorithm \ref{ChVB}. Regarding the score matching for $h$ and $\eta$, one can use the set of residuals of hVB obtained on the whole observation matrix, because the empirical estimation of $\mathcal{H}$-score (\ref{equ:corrmodel7}) will become more precise with a larger number of samples.

\begin{algorithm}[h]
	\caption{\textit{ChVB for Robust Source Imaging}}
	\label{ChVB}
	\begin{algorithmic}[1]		
		\State \textbf{input}:
		
		Observation matrix $\boldsymbol{B}\in\mathbb{R}^{M\times T}$;
		
		Leadfield matrix $\boldsymbol{G}\in\mathbb{R}^{M\times N}$;
		
		Prior information $a_{n0}$ $\&$ Prior weight $w$;
		
		\State \textbf{initialize}:
		
		Utilize hVB algorithm to obtain a preliminary source estimation $\hat{\boldsymbol{J}}_{hVB}\in\mathbb{R}^{N\times T}$ and then compute the residuals by $\boldsymbol{E}_{hVB}=\boldsymbol{B}-\boldsymbol{G}\hat{\boldsymbol{J}}_{hVB}$;
		
		For each observation channel: $m=1,...,M$, estimate the hyperparameters $h_m$ and $\eta_m$ using score matching (\ref{equ:corrmodel7}) with the residuals of $m$-th channel, i.e. $m$-th row of $\boldsymbol{E}_{hVB}$;
		
		\Repeat 
		\State $\boldsymbol{J}$\textit{-step}: update $\boldsymbol{J}$ according to (\ref{equ:chvb10});
		\State $\boldsymbol{A}$\textit{-step}: update $\boldsymbol{A}$ according to (\ref{equ:chvb13});
		\Until the reduction of free energy (\ref{equ:chvb2}) is sufficiently small or the number of iterations exceeds a predetermined limit;
		
		\State \textbf{output}:
		
		Source estimation $\hat{\boldsymbol{J}}_{ChVB}\in\mathbb{R}^{N\times T}$.
	\end{algorithmic}
\end{algorithm}

\section{Performance Evaluation}
\label{sec:evaluation}
To realize robust Bayesian source imaging for non-Gaussian sensor noise, this study proposed an improper likelihood model utilizing the MCC-based noise assumption $\mathcal{C}(e\mid0,h,\eta)$ which was integrated with the hierarchical prior distributions defined in hVB. Thus the crucial performance evaluation is to compare the proposed ChVB with hVB, both utilizing the identical prior distribution while different likelihood models, i.e. the Gaussian or the proposed one. Note that one can certainly utilize another prior distribution or optimization approach with our likelihood model, e.g. the Champagne family \cite{wipf2010robust,cai2018hierarchical,cai2019robust,cai2022bayesian}, which is further discussed in Section \ref{sec:dis}. A comparison between hVB and other source imaging methods was discussed in \cite{suzuki2021meg,takeda2019meg}.

\subsection{Simulation Study}
\label{sec:eval1}
For performance comparison between hVB and ChVB, first, this paper utilized a simulation study with known ground-truth of source activity, which can facilitate a quantitative evaluation by performance indicator. Our simulation study aims to realize experimental settings as close to the real-world source imaging tasks as possible. To this end, the source estimation result from a real-world dataset was used as the ground-truth for the source activity, while the reconstruction residuals were used as noises.

Specifically, we used Sub-01 from the neuroimaging dataset \cite{wakeman2015multi} (\href{https://openneuro.org/datasets/ds000117}{\textcolor{IEEEBlue}{\textit{https://openneuro.org/datasets/ds000117}}}) concerning the visual perception task with the stimulus of `Face'$/$`Scrambled' and the VBMEG toolbox (\href{https://vbmeg.atr.jp/v30}{\textcolor{IEEEBlue}{\textit{https://vbmeg.atr.jp/v30}}}) to generate the ground-truth for the simulation. The basic settings followed the toolbox tutorial \cite{takeda2019meg}. Notably, the simultaneously recorded 306-channel MEG and 70-channel EEG observation data were bandpass filtered between 1 and 40 Hz, and then downsampled to 100 Hz. The T1-MRI image was used to construct a polygon model for the cortical surface with 2,500 vertices, based on the coordinate from standard brain MNI-ICBM152 and FreeSurfer (\href{https://surfer.nmr.mgh.harvard.edu}{\textcolor{IEEEBlue}{\textit{https://surfer.nmr.mgh.harvard.edu}}}). Then the leadfield matrix was computed by a boundary element method based on 1-shell and 3-shell conductivity model of MEG and EEG, respectively. The \textit{t}-value of fMRI was employed as prior information, while the prior weight of fMRI information was 0.03. Based on these settings, the periods for 0 $-$ 0.49 sec (with 50 samplings) after the trigger for a total of 887 trials were used for source imaging which include 590 trials of `Face' visual stimuli and 297 trials of `Scrambled' visual stimuli, based on MEG and EEG signals with their leadfield and hVB algorithm. Afterwards, the results for the first twenty `Face' trials were selected and concatenated to serve as the true source activity, resulting in 1,000 samplings (50 samplings $\times$ 20 trials). Among 2,500 sources, only the top 30 with the largest activations were retained. The other sources were set to zeros to validate the localization correctness. Thus, the ground-truth source matrix $\boldsymbol{J}^*\in\mathbb{R}^{2500\times 1000}$ was obtained with 30 activated current dipoles. Accordingly, the pure sensor observations were acquired by multiplying the leadfield matrix with $\boldsymbol{J}^*$. To contaminate the observations using the real-world noise distribution, we utilized the reconstruction residuals from all 887 trials as the noise repository, because we perceived that the distribution of the reconstruction residuals was very heavy- tailed for both MEG and EEG, which potentially resulted from the infrequent physiological artifact. After the pure observation data were obtained, each element in the observation matrix was corrupted by a noise that was randomly sampled from the noise repository within the corresponding sensor channel. Moreover, to realize different levels for this corruption, the signal-to-noise ratio (SNR) was employed, defined as $\textit{SNR}=10\log_{10}\frac{\sigma^2_{Signal}}{\sigma^2_{Noise}}$, in which the numerator and denominator represent the variance for signal and noise, respectively. The original channel-average SNR for the pure observation data and the noise repository was $11.98$ and $-8.51$ for MEG and EEG, respectively. Accordingly we used several neighboring SNR values by adjusting the noise amplitude. To be specific, we employed $\text{SNR}=20,15,10,5,0$ for MEG, and $\text{SNR}=0,-5,-10,-15,-20$ for EEG. We also utilized the original noise sampled from the repository without tuning the amplitudes, denoted by Default SNR. The corrupted observations were utilized for source imaging without the prior information.

After obtaining the source estimations using hVB and ChVB with the above simulation setting, we assessed the performance by two typical quantitative indicators: aggregate and root mean square error (RMSE), defined as follows:
\begin{equation}
	\begin{split}
		\label{equ:eval1}
		Aggregate\triangleq&\frac{sCorr(\hat{\boldsymbol{J}},\boldsymbol{J}^*)+tCorr(\hat{\boldsymbol{J}},\boldsymbol{J}^*)}{2}\; \\
		RMSE\triangleq&\sqrt{\frac{1}{NT}\sum_{n=1}^{N}\sum_{t=1}^{T}\left(\hat{\boldsymbol{J}}_{n,t}-\boldsymbol{J}^*_{n,t}\right)^2}\; \\
	\end{split}
\end{equation}
in which the aggregate denotes the average between the spatial correlation $sCorr$ and temporal correlation $tCorr$. The spatial correlation is obtained by $\frac{1}{T}\sum_{t=1}^{T}corrcoef(|\hat{\boldsymbol{J}}_{t}|,|\boldsymbol{J}^*_{t}|)$ which denotes the averaged correlation coefficient between the spatial maps of the ground-truth and estimation along each time point. Temporal correlation is given by $\frac{1}{30}\sum_{n^*}corrcoef(\hat{\boldsymbol{J}}_{n^*},\boldsymbol{J}^*_{n^*})$, which is the average correlation coefficient of temporal domain for the 30 activated sources, and $n^*$ denotes the index for these activated current sources. The whole procedures for simulation were conducted for 50 repetitions to obtain generalized results.

\subsection{Real-World Dataset}
\label{sec:eval2}

In addition, we also utilized the above-mentioned real-world dataset \cite{wakeman2015multi} to directly evaluate the proposed ChVB algorithm. All the 16 subjects in this dataset were analyzed with a subject-specific brain model based on the T1-MRI image, in which the number of vertices was increased to 5,000 for a more elaborate modeling of cortical surface. We preprocessed the MEG signal utilizing the same method as Section \ref{sec:eval1}, and then estimated the source activity for the period of 0$-$0.49 sec after the visual stimulus concerning the first 100 `Face' trials for each subject. The prior weight for the subject-specific fMRI information was set as 0.03. hVB and ChVB were utilized for source estimation with these settings to investigate the neural activities regarding the perception for facial images based on the MEG recordings.

Finally, to objectively evaluate which algorithm could better estimate the task-related source activity, we used a downstream classification task based on the source imaging results obtained by hVB or ChVB. For each of all 16 subjects of the real-world dataset \cite{wakeman2015multi}, the periods for 0 $-$ 0.49 sec after the visual trigger for the first 100 `Face' trials and the first 100 `Scrambled' trials were concatenated for source imaging. After the sources were estimated by hVB and ChVB, we randomly selected $70\%$ data (70 `Face' trials and 70 `Scrambled' trials) and employed each trial's source activities (estimated by either hVB or ChVB) as the covariates to train a classification model regarding the label `Face'$/$`Scrambled'. Then, the classification model was tested on the remaining 60 trials. The classification performance was evaluated by the accuracy, which denotes the ratio of correctly classified testing samples. Because this was an extremely high-dimensional task, in which the number of covariates was 5,000 vertices $\times$ 50 samplings $=$ 250,000, we used a sparse classifier SLR \cite{yamashita2008sparse} that would automatically select the relevant features. Note that, the trials for `Face' and `Scrambled' were combined together in source estimation. Therefore, the label information would not be leaked during the source imaging processes. The random separation for the training$/$testing trials was conducted for 100 repetitions, based on the estimated source activity from MEG and EEG, respectively.

\section{Results}
\label{sec:res}

\subsection{Simulation Study}
\label{sec:res1}

\begin{figure}[t]
	\centering
	\includegraphics[width=0.86\columnwidth]{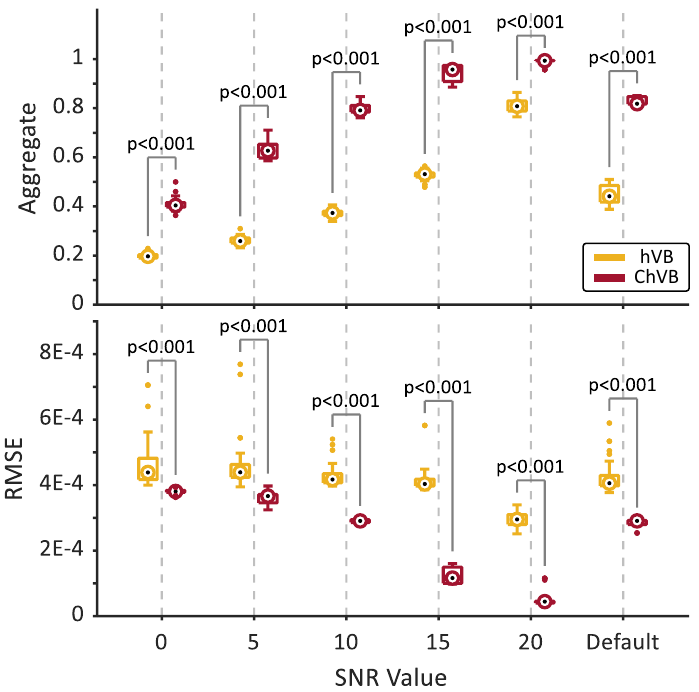}
	\caption{Simulation performance of MEG-based source imaging for different SNR values. Each boxplot contains 50 repetitions, and paired $t$-test is utilized to examine the statistical difference.}
	\label{FigSimResMeg}
\end{figure}

\begin{figure}[t!]
	\centering
	\includegraphics[width=0.86\columnwidth]{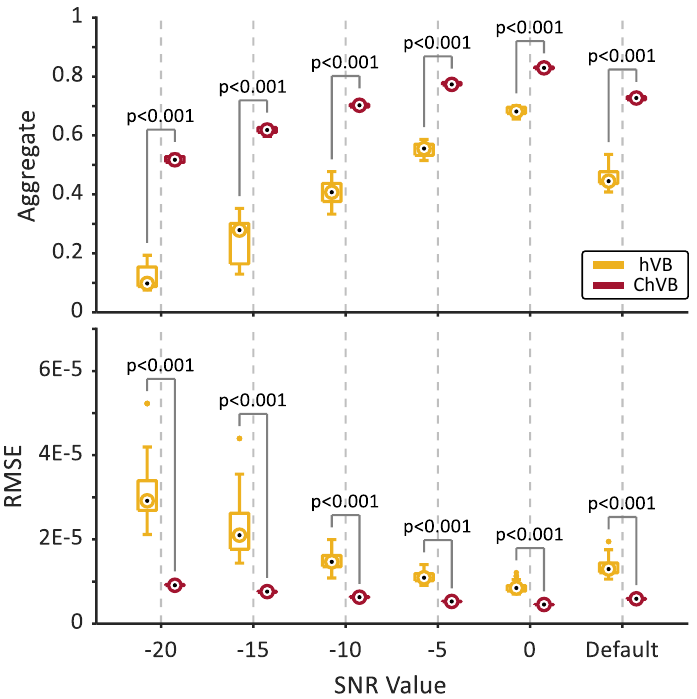}
	\caption{Simulation performance of EEG-based source imaging for different SNR values. Each boxplot contains 50 repetitions, and paired $t$-test is utilized to examine the statistical difference.}
	\label{FigSimResEeg}
\end{figure}

Fig. \ref{FigSimResMeg} and Fig. \ref{FigSimResEeg} show the simulation performance regarding MEG-based and EEG-based source imaging, respectively. One can perceive that, the proposed ChVB algorithm outperformed the conventional hVB on each SNR value for both MEG-based and EEG-based source imaging. Moreover, according to paired $t$-test, the improvements by ChVB were eminently statistically significant with small $p$-values ($p<0.001$ for each condition). In particular, considering the default SNR, the proposed ChVB increased the aggregate from $0.4477\pm0.0356$ (mean$\pm$standard deviation) to $0.8291\pm0.0153$ for MEG-based source imaging, and from $0.4539\pm0.0298$ to $0.7265\pm0.0072$ for EEG source imaging. The RMSE was decreased from $(4.21\pm0.43)\times10^{-4}$ to $(2.86\pm0.09)\times10^{-4}$ for MEG, and from $(1.34\pm0.19)\times10^{-5}$ to $(5.91\pm0.15)\times10^{-6}$ for EEG. This simulation demonstrates the superiority of ChVB for source imaging with non-Gaussian noises.

\subsection{Real-World Dataset}
\label{sec:res2}

\begin{figure*}[t!]
	\centering
	\includegraphics[width=1\textwidth]{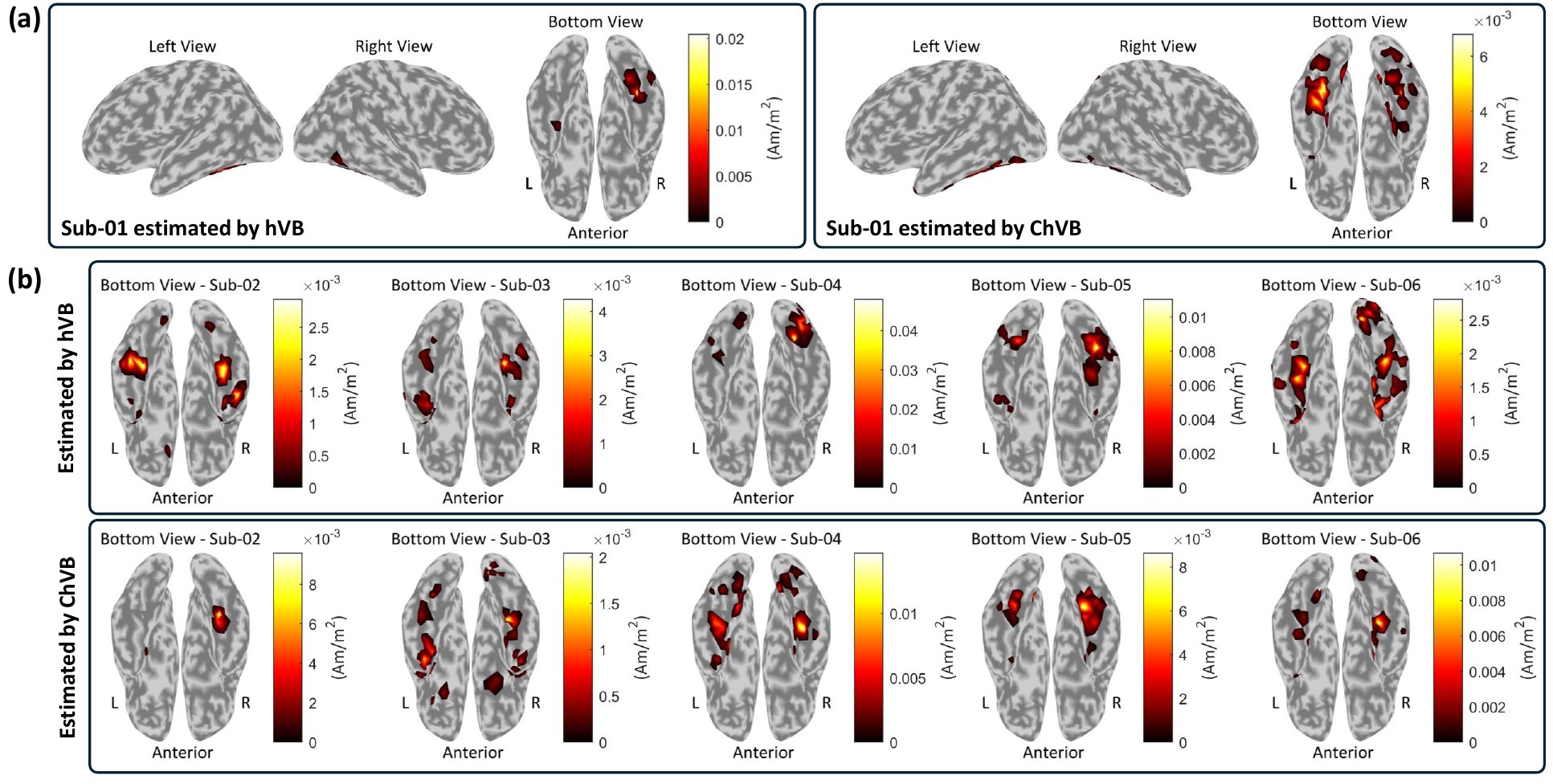}
	\caption{Reconstructed current map for the real-world neuroimaging dataset regarding the visual perception task with the stimulus of facial images. The color represents the absolute value of estimated currents for each source averaged across 0 $-$ 0.3 sec after the visual stimulus and averaged across 100 trials. The colored region denotes a larger activation than 0.1 times the maximum activation of all 5,000 estimated sources. (a) presents three views for the current map of Sub-01 (left panel: estimated by hVB, right panel: estimated by ChVB). (b) illustrates the bottom view for Sub-02$-$Sub-06 (upper panel: hVB, bottom panel: ChVB).}
	\label{FigRealMap}
\end{figure*}

\begin{figure}[htp!]
	\centering
	\includegraphics[width=0.96\columnwidth]{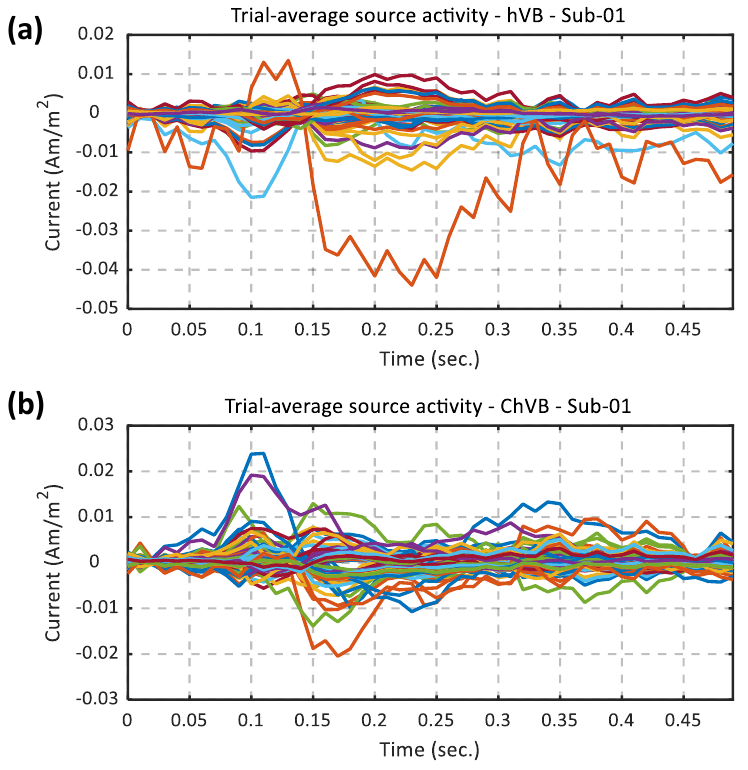}
	\caption{Reconstructed current time series for Sub-01. Each line illustrates the estimated current for a specific source. Since the majority of sources are not active, only those with values larger than 0.1 times the global maximum are displayed. (a) presents the estimation by hVB and (b) shows the estimation by ChVB. The illustrated time series are averaged by 100 trials with Face stimulus.}
	\label{FigRealTime}
\end{figure}  

Fig. \ref{FigRealMap} illustrates the reconstructed current map for 6 subjects (Sub-01$-$Sub-06) of the real-world neuroimaging dataset \cite{wakeman2015multi} regarding the visual perception task with the stimulus of facial images. Fig. \ref{FigRealMap}(a) shows the current map for Sub-01 from three angles. One could observe that, for both estimations from hVB and ChVB, almost all the activation regions were concentrated at the bottom view. In particular, for Sub-01 estimated by hVB (Fig. \ref{FigRealMap}(a) - Left), the right Fusiform Face Area (FFA) exhibited the most significant activation. By comparison, for ChVB (Fig. \ref{FigRealMap}(a) - Right), both left FFA and right FFA showed comparable activation for the face perception task. Previous studies pointed out that both left and right FFA are important for face detection and cognition \cite{wakeman2015multi,grill2004fusiform,grill2017functional}. Therefore,
ChVB realized more rational estimation for Sub-01 according to prior physiological knowledge. Fig. \ref{FigRealMap}(b) shows the maps for Sub-02$-$Sub-06 with the bottom view. One can observe that, almost every estimation located significant activation on left FFA or right FFA (or both) except for Sub-04 by hVB (Fig. \ref{FigRealMap}(b) - Upper). This estimation located high activation on the right Occipital Face Area (OFA), which is also responsible for the face perceptions. On the other hand, the estimation for Sub-02 by ChVB (Fig. \ref{FigRealMap}(b) - Bottom) may seem overly sparse, which only showed activation on right FFA. Since we cannot know the ground-truths for each subject in the real-world dataset, it could be difficult to objectively tell which current map is better.

\begin{figure*}[!htp]
	\centering
	\includegraphics[width=1\textwidth]{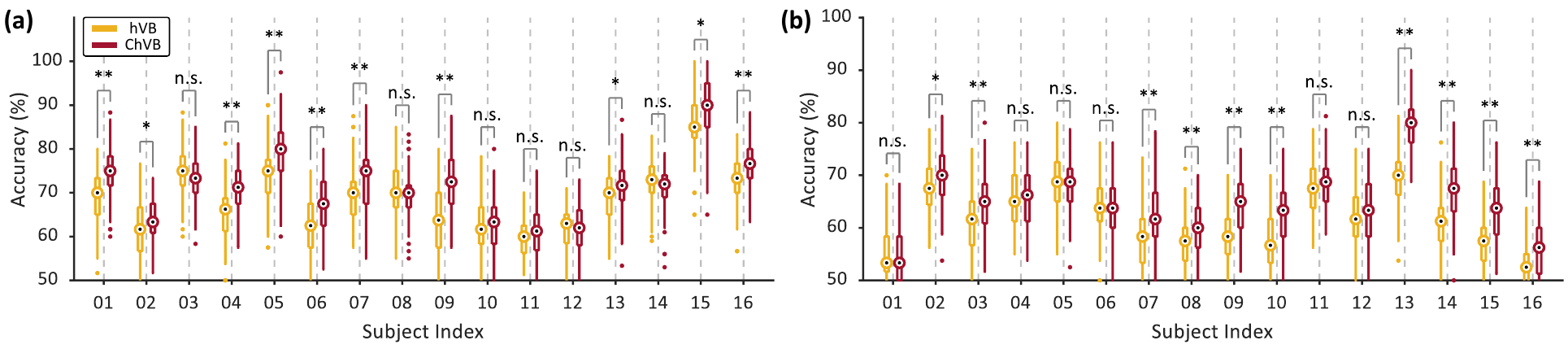}
	\caption{Classification accuracy for the downstream binary classification task between the `Face' and `Scrambled' stimuli utilizing the source activities estimated by either hVB or ChVB as the covariates. Each boxplot contains 100 repetitions with the random separation for training$/$testing trials. (a) utilizes the source estimation based on the MEG recordings, and (b) is obtained using the EEG recordings. Paired $t$-test is employed to examine the statistical difference ($**$: $p<0.001$, $*$: $p<0.05$, n.s.:$p>0.05$).}
	\label{FigResCls}
\end{figure*}

Fig. \ref{FigRealTime} presents the estimated trial-average current time series for Sub-01. One can perceive that the largest activation for the source estimation of hVB happened at 0.23 sec after the visual stimulus (Fig. \ref{FigRealTime}(a)). By comparison, in the estimation obtained by ChVB (Fig. \ref{FigRealTime}(b)), the largest activation happened at 0.1 sec and 0.17 sec after the visual trigger. Although we do not know the ground-truth for the time series, in our previous study \cite{takeda2019meg}, we conducted the source estimation for the same task utilizing denoised MEG signals by regressing out the electrooculogram (EOG) component, while note that in the present study we did not denoise the MEG signals with large efforts. We found that the result by ChVB utilizing noisy MEG recording (Fig. \ref{FigRealTime}(b)) is closer to that obtained by denoised MEG (Figure 6 in \cite{takeda2019meg}) than that by hVB (Fig. \ref{FigRealTime}(a)). This suggests that, even utilizing noisy observations, the proposed ChVB algorithm can provide the estimation similar to that obtained from less-noisy signals, which demonstrates the effectiveness of ChVB for suppressing the negative effect of non-Gaussian noises.

Considering the downstream classification between the Face and Scrambled visual stimuli, Fig. \ref{FigResCls} presents the classification accuracy for all 16 subjects based on MEG and EEG recording, respectively. One could find that, for the source estimation with MEG (Fig. \ref{FigResCls}(a)), ChVB realized higher classification accuracy than hVB with a significant difference ($p<0.05$, paired $t$-test) for 10 subjects. No significant difference was perceived for the other 6 subjects. For the EEG-based estimation, ChVB showed statistically higher accuracy than hVB in 10 subjects similarly, while no significant difference was observed for the remaining 6 subjects. This downstream classification result indicates that, the proposed ChVB is more likely to extract the current source activities which are relevant to the occurrent cognition process.

\section{Discussion}
\label{sec:dis}

To solve the potentially non-Gaussian observation noises for the Bayesian source imaging task, this study proposed a novel noise assumption motivated by the robust MCC method. Since our proposal essentially serves as a backbone for the Bayesian framework, i.e. the noise assumption, or equally, the likelihood function, our proposed method can be naturally combined with a wide variety of existing Bayesian source imaging approaches by replacing the non-robust Gaussian likelihood. For example, the automatic relevance determination (ARD) prior distribution utilized in the proposed ChVB algorithm may potentially result in the overly-sparse source estimation \cite{suzuki2021meg,takeda2019meg}. Accordingly, one could address this concern by leveraging the more rational prior distributions on the source activity, as investigated in the previous studies \cite{wipf2010robust,mohseni2014non,costa2017bayesian,cai2018hierarchical,cai2019robust,cai2023bayesian}. On the other hand, ChVB leveraged multiple approximations to compute the MAP estimations, such as the variational inference and the Laplacian approximation. Thus, there is a risk that the optimization might fail to obtain the satisfactory estimation. Therefore, one future study is to derive better optimization methods for the proposed likelihood function and the resultant posterior distribution with better guarantees.

Another noteworthy issue is the selection of hyperparameter for the correntropy. This study innovatively employed the score matching method to determine the hyperparameters of the new noise assumption, i.e. $h$ and $\eta$, which have a close relation with the kernel width $h_c$ for the original correntropy with $h_c=h/\eta$. Because score matching technique is designed as an alternative of maximum likelihood estimation to calculate the distribution parameter, it would necessitate the distributional interpretation for the model to be determined. Therefore, the direct utilization of score matching for the original MCC is impracticable, while the MCC-motivated noise assumption could well employ score matching technique to determine the hyperparameter, which is exactly one of the key contributions of this study. This provides an effective way to select the hyperparameter for MCC-related algorithm, in particular for the unsupervised learning task, e.g., the source imaging. Concerning score matching, we employed the reconstruction residuals from Gaussian model as the noises which is only an empirical method. In the future work, we will investigate a better approach to determine the hyperparameters with more theoretical guarantees.

Finally, we would also like to discuss some other limitations of this study and the corresponding future works. First, the new noise assumption $\mathcal{C}(e\mid0,h,\eta)$ exhibits a highly similar shape as the traditional Gaussian distribution except for the improper tails, as illustrated in Fig. \ref{FigMccCurve}. This motivates us to suppose that, the proposed noise assumption $\mathcal{C}(e\mid0,h,\eta)$ could be possibly unified to one distribution family with the traditional Gaussian model. Further properties for our proposed model $\mathcal{C}(e\mid0,h,\eta)$ would be investigated in the future studies. Second, we can see from Fig. \ref{FigMccCurve} that, the proposed noise assumption $\mathcal{C}$ is obviously a single-peak distribution with the peak at the origin. However, this single-peak property is not necessarily the ground-truth for brain recording noises. Previous studies showed that the multi-peak model can be more adequate for brain analysis \cite{li2021restricted,chen2018common}. Hence, it is worth investigating a multi-peak noise assumption in Bayesian regime for the future studies. Finally, the proposed noise assumption $\mathcal{C}(e\mid0,h,\eta)$ is currently a univariate version which cannot take the interaction into consideration. However, the real-world noises would interact across recording channels. Thus, the proposed $\mathcal{C}(e\mid0,h,\eta)$ would be less adequate when the recording noises highly correlate between the measurement channels. Our future works will also concentrate on developing a multivariate version for the proposed noise assumption which may further improve the performance for the real-world source imaging applications.

\section{Conclusion}
\label{sec:con}

This paper proposed a novel robust Bayesian source imaging algorithm called ChVB. In particular, we derived a novel noise assumption motivated by the robust MCC method, and utilized it to structure the likelihood function. The experimental results on both simulation and the real-world dataset demonstrated the effectiveness of ChVB for eliminating the adverse effects from non-Gaussian observation noises on Bayesian source imaging. Our proposal which plays the role of backbone in the Bayesian learning framework could be naturally combined with existing source imaging approaches for further improvements. We also provided insightful discussions, including some future studies. The proposed ChVB algorithm will be also incorporated in our developed VBMEG toolbox (\href{https://vbmeg.atr.jp}{\textcolor{IEEEBlue}{\textit{https://vbmeg.atr.jp}}}) in the future release to facilitate the real-world applications for the ESI task.

\bibliographystyle{IEEEtran}
\bibliography{BiBib}
\end{document}